\documentclass[a4paper,10pt]{article} % 基本フォントサイズは10ptに設定
\usepackage{times}
\usepackage[colorlinks=true, urlcolor=blue, citecolor=black]{hyperref}
\usepackage[left=20mm,right=20mm,top=25mm,bottom=24mm,columnsep=8mm]{geometry}
\usepackage{multicol}
\usepackage{caption}
\usepackage{subcaption}
\usepackage{wrapfig}
\usepackage[absolute,overlay]{textpos}
\usepackage{titlesec}
\usepackage[backend=bibtex,style=numeric, sorting=none]{biblatex} % 参考文献の管理用パッケージ
\usepackage{titling}
\titleformat{\section}{\fontsize{10pt}{12pt}\bfseries}{\thesection.}{1em}{}
\titleformat{\subsection}{\fontsize{10pt}{12pt}\bfseries}{\thesubsection.}{1em}{}

\titlespacing*{\subsection}{0pt}{*1}{*0}

\usepackage{amsmath}
\usepackage{amssymb}
\usepackage{mathtools}
\usepackage{amsthm}
\allowdisplaybreaks

\begin{document}

% Display "ISASE 2024" at the top left, aligned to the left margin
\title{{\fontsize{16pt}{16pt}\selectfont \textbf{Off-Policy Evaluation for Recommendations} \\ \textbf{with Missing-Not-At-Random Rewards}}}
% ISASE 2024をページ上部から25mmの位置で左寄せ
\begin{textblock*}{20mm}(15mm, 25mm) % 左上の位置をtop=25mmに調整
    \normalsize \textbf{ISASE 2025}
\end{textblock*}
\date{}

\maketitle

\vspace{-2.7cm}

% Author(s): 12 Point Normal / Centering
\begin{center}
    {\fontsize{12pt}{12pt}\selectfont Tatsuki Takahashi*, Chihiro Maru** and Hiroko Shoji**}
\end{center}

% Affiliation(s) and mailing address(es): 9 Point Italic / Centering
\begin{center}
    {\fontsize{9pt}{11pt}\selectfont 
    \textit{*Chuo University Graduate School, 1-13-27 Kasuga, Bunkyo Ward, Tokyo, Japan}\\
    \textit{a23.kb3g@g.chuo-u.ac.jp}\\
    \textit{**Chuo University, 1-13-27 Kasuga, Bunkyo Ward, Tokyo, Japan}\\
    \textit{cmaru671@g.chuo-u.ac.jp, hiroko@kc.chuo-u.ac.jp}\\
    }
\end{center}

\vspace{1em} % スペース調整

% Abstract: 9 Point Normal / Single Column & Fully Justified
{\noindent \fontsize{9pt}{9pt}\selectfont
\textbf{Abstract:} Unbiased recommender learning (URL) and off-policy evaluation/learning (OPE/L) techniques are effective in addressing the data bias caused by display position and logging policies, thereby consistently improving the performance of recommendations. However, when both bias exits in the logged data, these estimators may suffer from significant bias. In this study, we first analyze the position bias of the OPE estimator when rewards are missing not at random. To mitigate both biases, we propose a novel estimator that leverages two probabilities of logging policies and reward observations as propensity scores. Our experiments demonstrate that the proposed estimator achieves superior performance compared to other estimators, even as the levels of bias in reward observations increases.
}

%\vspace{1em} % スペース調整

% Keywords: 9 Point Italic / Single Column (aligned with left margins)
{\noindent \fontsize{9pt}{9pt}\selectfont \textbf{Keywords:} \textit{Recommender System, Off-Policy Evaluation/Learning, Unbiased Recommender Learning, Inverse Propensity Score}}

\begin{multicols}{2}

% Section and subsection headings: 10 Point Bold / Double Column / Fully Justified

\section{INTRODUCTION}

Recommender systems are widely employed to assist individuals in making decisions within the web industry. Conducting offline assessments prior to deployment is essential to continuously enhance the performance of recommender policies. This approach helps avoid the high cost A/B testing and ensures that relevant items are presented to each user \cite{saito2020open,chen2019top,gruson2019offline}. Furthermore, it is crucial to consider the data bias introduced by the policies and display positions in offline assessments \cite{chen2023bias}. This measure prevents the algorithm from over-fitting items (actions) that were easily observable in the logged data.

Modeling the existence of counterfactuals yields fair recommendations and enhances the predictability of the performance of an algorithm. To address the position bias in recommended displays, we can utilize unbiased recommender learning (URL) techniques that leverage the probabilities of reward observations for each user as inverse propensity scores (IPS) for training \cite{schnabel2016recommendations,saito2020unbiased}. Specifically, the loss functions used in the learning-to-rank context, such as pointwise, are proposed with built-in propensity score \cite{saito2020unbiased}. To address the biases inherent in logging policies, off-policy evaluation/learning (OPE/L) techniques that leverage action (item) selection probabilities of logging policies as IPS can be employed for the estimation of a new policy \cite{swaminathan2015counterfactual,saito2022off,li2018offline}. However, while IPS does not introduce bias, it suffers from high variance issues owing to its importance weights as the action space expands \cite{saito2022off,li2018offline}. Therefore, it is essential to consider the trade-off between bias and variance to estimate and learn policy values accurately. To address the high variance of the IPS, the marginalized IPS (MIPS) utilizes marginalized importance weights over action embedding spaces. Under
\noindent
\begin{minipage}{\linewidth}
    \centering
    \includegraphics[width=\linewidth]{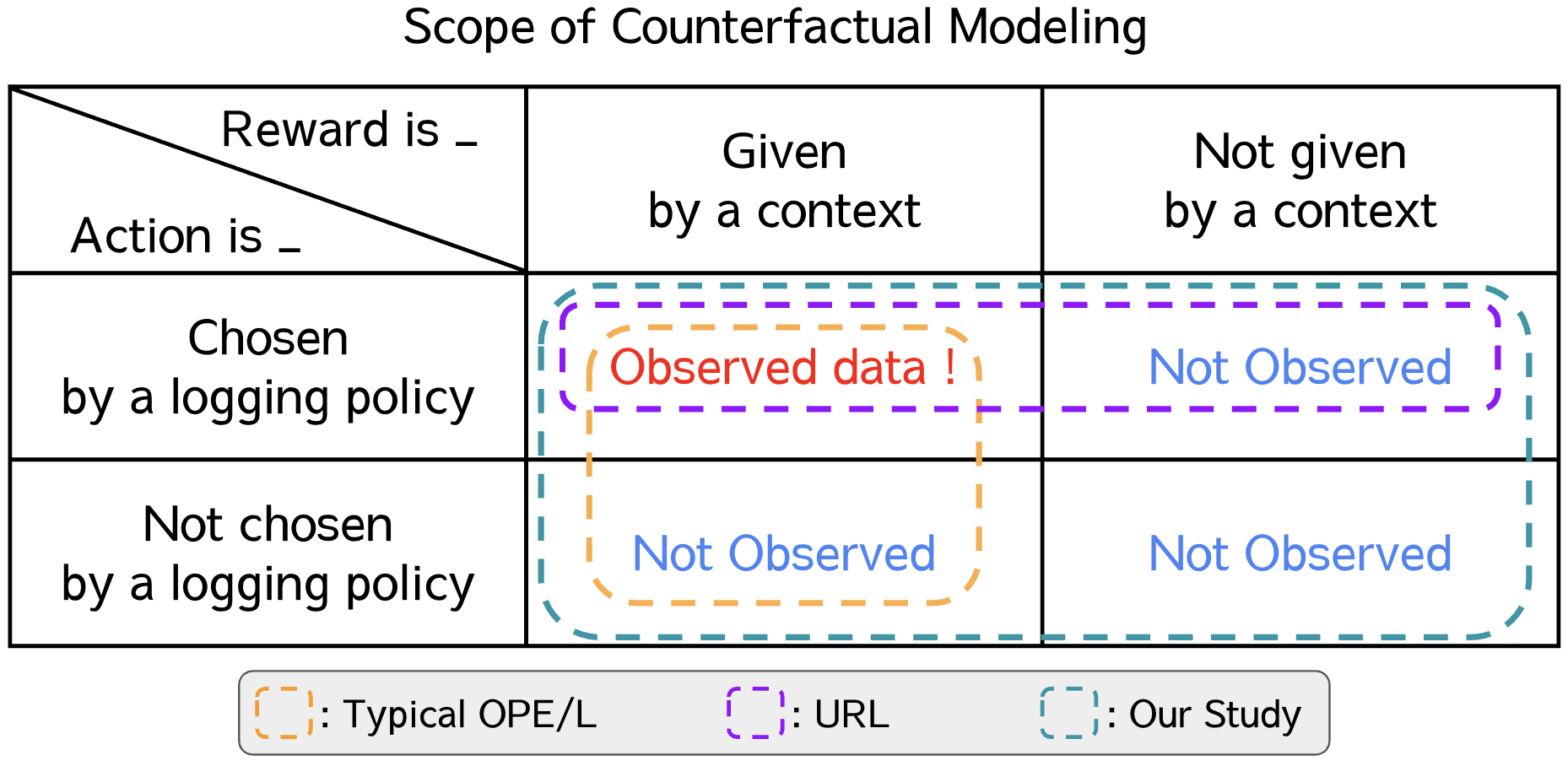}
    \captionof{figure}{Scope of counterfactual modeling in our study}
    \label{fig:scope_counterfactuals}
\end{minipage}
\vspace{1mm}

\noindent
the assumption that there is no direct effect, meaning the action is not influenced by the reward function, the MIPS avoids introducing bias while simultaneously reducing variance compared to IPS \cite{saito2022off}. 

Although the URL and OPE/L can mitigate position bias and the biases associated with logging policies, if both display position and logging policy biases are present in the logged data, which is a realistic scenario, these techniques may encounter significant bias issues. The IPS in the pointwise loss function in URL can be considered a variant of the direct method (DM) in OPE. Consequently, this IPS is recognized for introducing the biases due to prediction errors while not introducing position bias \cite{saito2022off}. In this study, we first demonstrate that when rewards are missing not at random (MNAR) owing to position bias, OPE estimators introduce the bias, although they do not introduce the bias associated with logging policies. We then model counterfactuals with respect to both action and reward observations, as shown in Figure 1. We propose a novel OPE estimator, \textit{a MIPS with reward observation IPS (MIPS (w/ROIPS))}, which leverages the embedding selection probabilities of logging policies and the probabilities of reward observations for each position as propensity scores. We demonstrate that the MIPS (w/ROIPS) is unbiased under several assumptions, even in the presence of potential biases from logging policies and display position. Finally, in the synthetic experiments, we demonstrate that although we cannot obtain the true probabilities of reward observations, which is unknown in realistic scenario, we can balance the trade-off between bias and variance to effectively utilize the heuristic ROIPS, which we can estimate using only logged data, even in the presence of both biases.

\section{BACKGROUND AND RELATED WORK}
Here, we formulate the URL and OPE/L.

\subsection{Unbiased recommender learning}
URL, particularly when employing a pointwise loss function, aims to learn the relevance of each user-item pair using only biased data that arises from display position. We denote \( u \in \mathcal{U} \) and \( i \in \mathcal{I} \) as the indices for users and items, respectively. Additionally, we define random variables relevance \( R_{u,i} \in \mathbb{R} \) and observation \( O_{u,i} \in \{ 0, 1 \} \), each associated with parameters \( \gamma_{u,i} \) and \( \theta_{u,i} \), respectively, for all user-item pairs \( (u,i) \). Consequently, we define the ideal loss function to be estimated as follows:

\vspace{-5mm}
\begin{align*}
    \mathcal{J}(f_{\phi}) := \sum_{u \in \mathcal{U},i \in \mathcal{I}} \delta(\gamma_{u,i}, f_{\phi}(u,i)).
\end{align*}
\vspace{-5mm}

where \( f_{\phi} \) is a recommendation algorithm, such as matrix factorization, and \( \delta(\cdot,\cdot) \) is a loss function, such as squared and cross-entropy errors. However, because we cannot access all pairs of \( \gamma_{u,i} \), we must utilize an estimator using only biased data. The IPS for addressing observation bias \cite{schnabel2016recommendations,saito2020unbiased} is defined as follows:

\vspace{-5mm}
\begin{align}
    \hat{J}_{\text{IPS}}(f_{\phi};\mathcal{D}) := \sum_{(u,i) \in \mathcal{D}} \frac{O_{u,i}}{\theta_{u,i}} \delta(R_{u,i}, f_{\phi}(u,i)).
\end{align}
\vspace{-3mm}

where \( \mathcal{D} \) is the observable logged data. The IPS utilizing \( \theta_{u,i} \) as a propensity score is unbiased for any \( f_{\phi} \) (i.e., \( \mathbb{E}_{\mathcal{D}}[\hat{\mathcal{J}}_{\text{IPS}}(f_{\phi};\mathcal{D})] = \mathcal{J}(f_{\phi}) \)).

\subsection{Off-policy evaluation/learning}
OPE aims to accurately estimate the value of a new policy using only logged data, which contains the logging policies biases. We first begin the notation. We define the user context, \( x \in \mathcal{X} \), generated from unknown distribution \( p(x) \) and the ranking action (item), \( \boldsymbol{a} := (\boldsymbol{a}_1,\cdots,\boldsymbol{a}_K) \in \Pi(\mathcal{A}) \), generated from the known distribution of logging policies \( \pi_{0}(\boldsymbol{a}|x) \) given \( x \). Here, \( K \) is the length of the ranking. We denote the deterministic action embedding (e.g., category) as \( \boldsymbol{e}_{\boldsymbol{a}} := (\boldsymbol{e}_{\boldsymbol{a}_1},\cdots,\boldsymbol{e}_{\boldsymbol{a}_K}) \in \Pi(\mathcal{E}) \).
We then define reward \( \boldsymbol{r} \) generated from unknown distribution \( p(\boldsymbol{r}|x,\boldsymbol{a}) \) given \( x \) and \( \boldsymbol{a} \). The value of target policy \( \pi \) to be estimated is defined as \( V(\pi) = \sum_{k=1}^{K} V^{(k)}(\pi) \), which is the sum of the position-wise values, as follows:

\vspace{-5mm}
\begin{align}
    V^{(k)}(\pi) = \mathbb{E}_{p(x)\pi(\boldsymbol{a}|x)}[q_{k}(x, \boldsymbol{a})].
\end{align}
\vspace{-5mm}

where \( q_{k}(x, \boldsymbol{a}) = \mathbb{E}[\boldsymbol{r}_k|x,\boldsymbol{a}] \) is the expected reward for each position \( k \) given \( x \) and \( \boldsymbol{a} \). We can obtain \( n \) logged data \( \mathcal{D} := \{ (x_i, \boldsymbol{a}_i, \boldsymbol{e}_{\boldsymbol{a}_i}, \boldsymbol{r}_i) \}_{i=1}^{n} \). This data is generated from \( (x,\boldsymbol{a}, \boldsymbol{e}_{\boldsymbol{a}}, \boldsymbol{r}) \sim p(x)\pi_{0}(\boldsymbol{a}|x)p(\boldsymbol{r}|x,\boldsymbol{a}) \), independently and identically.

We then define the position-wise values of the MIPS estimator \cite{saito2022off}, which addresses the logging policies bias, as follows:

\vspace{-6mm}
\begin{align}
    \hat{V}^{(k)}_{\text{MIPS}}(\pi;\mathcal{D}) := \frac{1}{n}\sum_{i=1}^{n} w(x_i, \boldsymbol{e}_{\boldsymbol{a}_{i,k}}) \boldsymbol{r}_{i,k}.
\end{align}
\vspace{-4mm}

where \( w(x, \boldsymbol{e}_{\boldsymbol{a}_k}) := \pi(\boldsymbol{e}_{\boldsymbol{a}_k}|x) / \pi_{0}(\boldsymbol{e}_{\boldsymbol{a}_k}|x) \) is the marginalized importance weight over action embedding spaces, and \( \pi(\boldsymbol{e}_{\boldsymbol{a}_k}|x) \) denotes the marginalized distribution that determines the action embedding for each position. In OPE, to evaluate the performance of the estimator, it is common practice to assess the mean squared error (MSE), which can be decomposed into the squared bias and estimator variance terms, as follows:

\vspace{-5mm}
\begin{align}
    \text{MSE}(\hat{V}(\pi;\mathcal{D})) &:= \mathbb{E}_{\mathcal{D}} \Big[(V(\pi) - \hat{V}(\pi;\mathcal{D}))^2 \Big], \\
    &= \text{Bias}(\hat{V}(\pi;\mathcal{D}))^2 + \mathbb{V}_{\mathcal{D}}[\hat{V}(\pi;\mathcal{D})]. \notag
\end{align}
\vspace{-5mm}

where \( \text{Bias}(\hat{V}(\pi;\mathcal{D})) := V(\pi) - \mathbb{E}_{\mathcal{D}}[\hat{V}(\pi;\mathcal{D})] \) is the difference between the true policy value and expected value of the estimator, and \( \mathbb{V}_{\mathcal{D}}[\hat{V}(\pi;\mathcal{D})] := \mathbb{E}_{\mathcal{D}}[(\hat{V}(\pi;\mathcal{D}) - \mathbb{E}_{\mathcal{D}}[\hat{V}(\pi;\mathcal{D})])^2] \) is the estimator variance. Therefore, to minimize the MSE, we should consider the trade-off between the bias and variance of the estimator. In contrast, OPL aims to identify the optimal policy, \( \pi_{\phi}^{*} \), parameterized by \( \phi \) using only logged data \cite{swaminathan2015counterfactual}.

The MIPS becomes unbiased (i.e., \( \mathbb{E}_{\mathcal{D}}[\hat{V}^{(k)}_{\text{MIPS}}(\pi;\mathcal{D})] = V^{(k)}(\pi) \)) under a common embedding support \cite{saito2022off} when the following assumption of no direct effects is satisfied. Furthermore, under these assumptions, the MIPS reduces the variance, as compared to the IPS \cite{saito2022off}.

\textbf{Assumption 1.} (\textit{no direct effect}) Rewards \( \boldsymbol{r} \) is not influenced by ranking actions \( \boldsymbol{a} \) (i.e., \( \mathbb{E}[\boldsymbol{r}_k|x,\boldsymbol{a}] = \mathbb{E}[\boldsymbol{r}_k|x,\boldsymbol{e}_{\boldsymbol{a}_k}] \) ).

\textbf{Limitation of previous studies} When considering the IPS(Eq.(1)) in URL as a variant of DM in the context of OPE that predicts the reward function directly, the IPS is subject to significant bias owing to its prediction errors. Similarly, OPE estimators may suffer from significant bias, although they do not introduce the bias associated with logging policies when rewards are MNAR, which is a realistic scenario. Therefore, it is desirable to develop a novel estimator that can effectively address both biases to recommend relevant items to each user.

\section{OUR PROPOSAL}
Here, we first demonstrate that the MIPS introduces a bias when rewards are MNAR owing to display positions. We then propose a novel estimator that combines URL and OPE techniques to address both biases associated with logging policies and reward observations. 

\subsection{Bias of OPE estimator when rewards are MNAR}
First, we define the random variable for reward observations \( \boldsymbol{o} = (\boldsymbol{o}_1, \cdots, \boldsymbol{o}_K ) \in \Pi(\mathcal{O}) \) generated from unknown distribution \( \theta(\boldsymbol{o}|x) \) given \( x \). Here, \( \mathcal{O} = \{ 0, 1 \} \) is the observation candidates, indicating that when only \( \boldsymbol{o}_k = 1 \), we can observe the instance of logged data for position \( k \). We then extend the data generation process to incorporate \( \boldsymbol{o} \) as follows:

\vspace{-5mm}
\begin{align*}
    \{(x, \boldsymbol{a}, \boldsymbol{e}_{\boldsymbol{a}}, \boldsymbol{o}, \boldsymbol{r}) | \boldsymbol{o}_k = 1 \} \sim p(x)\pi_{0}(\boldsymbol{a}|x)\theta(\boldsymbol{o}|x)p(\boldsymbol{r}|x,\boldsymbol{a}).
\end{align*}
\vspace{-5mm}

Therefore, we can obtain \( n \) logged data \( \mathcal{D} = \{(x_i, \boldsymbol{a}_i, \boldsymbol{e}_{\boldsymbol{a}_i}, \boldsymbol{o}_i, \boldsymbol{r}_i) | \boldsymbol{o}_{i,k} = 1, \forall k \}_{i=1}^{n} \) independently and identically. In this case, the MIPS has the following bias.

\textbf{Theorem} \textit{The MIPS has the following bias for each position, although Assumption 1 holds when rewards are MNAR.}

\vspace{-5mm}
\begin{multline*}
    \text{Bias}\Big(\hat{V}^{(k)}_{\text{MIPS}}(\pi; \mathcal{D})\Big) \\= \mathbb{E}_{p(x)\pi(\boldsymbol{e}_{\boldsymbol{a}_k}|x)} \Big[ q_{k}(x,\boldsymbol{e}_{\boldsymbol{a}_k}) \left( 1 - \theta(\boldsymbol{o}_k|x) \right) \Big].
\end{multline*}
\vspace{-5mm}

\begin{proof}
\vspace{-5mm}
\begin{flalign*}
    &\mathbb{E}_{\mathcal{D}}\Big[\hat{V}^{(k)}_{\text{MIPS}}(\pi;\mathcal{D})\Big] = \mathbb{E}_{\mathcal{D}} \left[ \frac{1}{n} \sum_{i=1: \boldsymbol{o}_{i, k} = 1}^{n} w(x_i,\boldsymbol{e}_{\boldsymbol{a}_{i,k}})\boldsymbol{r}_{i,k} \right] &\\
    &= \mathbb{E}_{p(x)\pi(\boldsymbol{e}_{\boldsymbol{a}_k}|x)\theta(\boldsymbol{o}|x)} \Big[q_{k}(x,\boldsymbol{e}_{\boldsymbol{a}_k}) \mathbb{I} \{ \boldsymbol{o}_{k} = 1 \}  \Big] 
    \substack{\because \text{i.i.d and} \\ \text{Assumption 1.}}
\end{flalign*}
\vspace{-5mm}

Therefore, we have the following bias:

\vspace{-5mm}
\begin{flalign*}
     &\text{Bias}\Big(\hat{V}^{(k)}_{\text{MIPS}}(\pi; \mathcal{D})\Big) = V^{(k)}(\pi) - \mathbb{E}_{\mathcal{D}} \Big[\hat{V}^{(k)}_{\text{MIPS}}(\pi;\mathcal{D})\Big] &\\
     &= \mathbb{E}_{p(x)\pi(\boldsymbol{e}_{\boldsymbol{a}_k}|x)} \Big[ q_{k}(x,\boldsymbol{e}_{\boldsymbol{a}_k}) \left( 1 - \theta(\boldsymbol{o}_k|x) \right) \Big]
\end{flalign*}
\vspace{-3mm}
\end{proof}

where \( \theta(\boldsymbol{o}_k|x) = \sum_{\boldsymbol{o}} \theta(\boldsymbol{o}|x) \mathbb{I} \{ \boldsymbol{o}_k = 1 \} \) is the marginal distribution of reward observations over observation spaces for each position given \( x \), and \( \mathbb{I}\{ \cdot \} \) is the indicator function. It returns 1 if the proposition within the function is true; otherwise, it returns 0. This theorem suggests that the MIPS rises a larger bias as \( \theta(\boldsymbol{o}_k|x) \) approaches zero for each position. Additionally, this theorem implies that all basic OPE/L estimators, including the MIPS, have a bias because they implicitly assume that all rewards can be observed without any omissions.

\subsection{MIPS with reward observation IPS }
To overcome both biases associated with logging policies and reward observations, we propose a novel estimator: \textit{MIPS (w/ROIPS)}, which leverages both the probability embedding selection and reward observations as propensity scores, as follows:

\vspace{-5mm}
\begin{align}
    \hat{V}^{(k)}_{\text{MIPS (w/ROIPS)}}(\pi; \mathcal{D}) := \frac{1}{n} \sum_{\substack{i=1 \\: \boldsymbol{o}_{i,k} = 1}}^{n} \frac{w(x_i, \boldsymbol{e}_{\boldsymbol{a}_{i,k}})\boldsymbol{r}_{i,k}}{\theta(\boldsymbol{o}_{k}|x_i)}.
\end{align}
\vspace{-5mm}

The MIPS (w/ROIPS) does not introduce the bias under Assumption 1, although rewards are MNAR.

\textbf{Proposition} \textit{MIPS (w/ROIPS) is unbiased under Assumption 1 (i.e., \( \mathbb{E}_{\mathcal{D}}[\hat{V}^{(k)}_{\text{MIPS (w/ROIPS)}}(\pi;\mathcal{D})] = V^{(k)}(\pi) \))}.

\vspace{-1mm}
\begin{proof}
\begin{flalign*}
    & \mathbb{E}_{\mathcal{D}}\Big[\hat{V}^{(k)}_{\text{MIPS (w/ROIPS)}}(\pi;\mathcal{D}) \Big] &\\
    &=  \mathbb{E}_{\mathcal{D}} \left[ \frac{1}{n} \sum_{i=1: \boldsymbol{o}_{i, k} = 1}^{n} w(x_i, \boldsymbol{e}_{\boldsymbol{a}_{i,k}}) \frac{\boldsymbol{r}_{i,k}}{\theta(\boldsymbol{o}_{k}|x_i)} \right] &\\
    &= \mathbb{E}_{p(x)\pi(\boldsymbol{e}_{\boldsymbol{a}_k}|x)} \left[q_{k}(x,\boldsymbol{e}_{\boldsymbol{a}_k}) \frac{ \sum_{\boldsymbol{o}^{\prime}} \theta(\boldsymbol{o}^{\prime}|x) \mathbb{I} \{ \boldsymbol{o}^{\prime}_{k} = 1 \}}{\theta(\boldsymbol{o}_k|x)}  \right] &\\
    &\qquad \because \text{i.i.d and Assumption 1.} \\
    &= V^{(k)}(\pi)
\end{flalign*}
\end{proof}
\vspace{-8mm}

\section{SYNTHETIC EXPERIMENTS}
Here, we conduct an experiment to perform the MIPS (w/ROIPS) as the bias level of reward observations varies. Synthetic experiments are effective for assessing the theoretical properties of the estimator \cite{saito2020unbiased,saito2022off}. Our experiment implementation is available at \href{https://github.com/tatsuki1107/isase2025-ope-mnar-reward}{https://github.com/tatsuki1107/isase2025-ope-mnar-reward}. 

\subsection{Setup} We generate the logged data based on the \textit{open bandit pipeline} \cite{saito2020open} and settings from a previous study \cite{saito2022off}. First, we create the five-dimensional contexts, \( x \), generated from a standard normal distribution. We then define a softmax policy, where the input values are the expected reward, \( q_{k}(x,\boldsymbol{e}_{\boldsymbol{a}_k}) \), which satisfies Assumption 1 as the logging policy. In this context, we set the number of actions \( |\mathcal{A}| \) to \(500\), number of action categories \( |\mathcal{E}|\) to\( 5 \), and length of rankings \( K \) to \( 5 \) as the default values. We then define the ranking action space as \( |\Pi(\mathcal{A})| = |\mathcal{A}| \times |\mathcal{E}|^{K-1}  \). Next, we synthesize the probabilities of reward observations \( \theta(\boldsymbol{o}|x) \), which decreases as the display position moves lower, similar to observations in the real world, as follows: 

\vspace{-5mm}
\begin{align*}
    \theta(\boldsymbol{o}|x) := \frac{|x^{\top} v_{\boldsymbol{o}}| \cdot \alpha^{K - \bar{\boldsymbol{o}}}}{\sum_{\boldsymbol{o}^{\prime} \in \Pi(\mathcal{O})} |x^{\top} v_{\boldsymbol{o}^{\prime}}| \cdot \alpha^{K - \bar{\boldsymbol{o}^{\prime}}}}.
\end{align*}
\vspace{-3mm}

where \( v_{\boldsymbol{o}} \) is the unknown parameter for each \( \boldsymbol{o} \) sampled from the standard normal distribution, \( \bar{\boldsymbol{o}} := \sum_{k=1}^{K} \boldsymbol{o}_k \) is the total number of observations, and \( \alpha \geq 0 \) is the parameter that determines the strength of the position bias. When \( \alpha = 0 \), no position bias exists; the position bias increases as \( \alpha \) increases. Finally, we observe rewards \( \boldsymbol{r}_k \) from the normal distribution, with \( q_{k}(x,\boldsymbol{e}_{\boldsymbol{a}_k}) \) serving as the parameters. To summarize, we obtain \( n \) logged data points \( \mathcal{D} = \{ (x_i, \boldsymbol{a}_i, \boldsymbol{e}_{\boldsymbol{a}_i}, \boldsymbol{o}_i, \boldsymbol{r}_i) | \boldsymbol{o}_{i,k} = 1 \}_{i=1}^{n} \) from the joint distribution \( p(x)\pi_{0}(\boldsymbol{a}|x)\theta(\boldsymbol{o}|x)p(\boldsymbol{r}_k|x,\boldsymbol{e}_{\boldsymbol{a}_k}) \). We set \( n = 1000 \) as the default value. We then define an epsilon-greedy policy, where input values are \( q_{k}(x,\boldsymbol{e}_{\boldsymbol{a}_k}) \), as the target policy.

\begin{figure*}[ht]
    \centering
    \includegraphics[width=\textwidth]{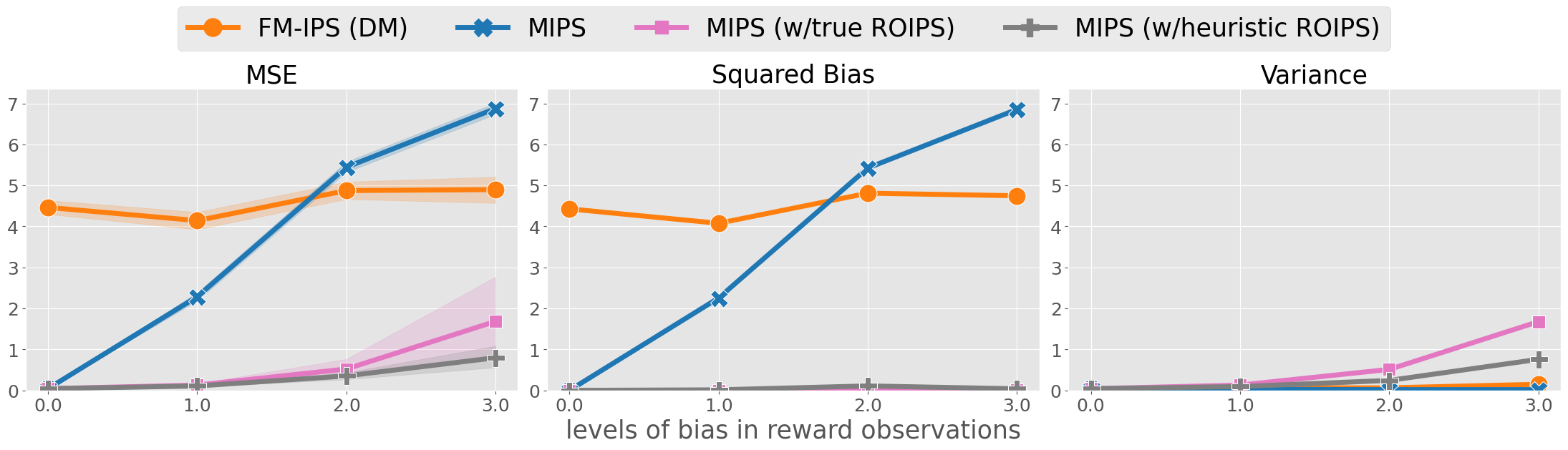}
    \vspace{-5mm}
    \caption{MSE, squared bias, and variance with \textbf{varying levels of bias in reward observations}. Note that, as shown in Eq.(4), the MSE (left) can be decomposed into the squared bias term (middle) and the variance term of the estimator (right).}
    \label{fig:images}
\end{figure*}

\subsection{Baselines} We utilized the following estimators in our experiments. \textit{Factorization machine (FM)-IPS (DM)}: This refers to the application of an FM to \( f_{\phi} \) in Eq.(1), which aims to eliminate position bias. We train by mapping \( R_{u,i} \) to \( \boldsymbol{r}_k \) and \( \theta_{u,i} \) to \( \theta(\boldsymbol{o}_k|x) \). \textit{MIPS}: This is the OPE estimator in Eq.(3), which aims to eliminate the bias associated with logging policies. \textit{MIPS(w/true ROIPS)}: This estimator is a fusion of the OPE and URL techniques in Eq.(5), which aims to eliminate both display position and logging policies biases. \textit{MIPS(w/heuristic ROIPS)}: This estimator utilizes the heuristic observation probabilities, which are estimated using the logged data. In practice, we cannot obtain the true observation probabilities. Specifically, we average observation \( \hat{\theta}(\boldsymbol{o}_k|\cdot) = \frac{1}{n} \sum_{i=1}^{n} \boldsymbol{o}_{i,k} \) for each position, which is context-free probability.

\subsection{Results}
We assess the aforementioned estimators using duplicated logged data with different random seeds \( \mathcal{S} \). We set \( |\mathcal{S}| = 100 \). Subsequently, we calculate the MSE value, as defined in Eq.(4), for each estimator.

\textbf{How does our estimator perform with varying levels of bias in reward observations?} We varied \( \alpha \in \{ 0.0, 1.0, 2.0, 3.0 \} \). The results are presented in Figure 2. We observe that the MIPS (w/heuristic ROIPS) achieved the lowest MSE, even as the position bias increased. Specifically, the MIPS (w/heuristic ROIPS) improved the MSE by approximately 88.4 \% compared to the MIPS when \( \alpha = 3.0 \).
This is because the MIPS (w/heuristic ROIPS) can reduce variance as its heuristic probabilities are simpler compared to the true value while introducing some biases. The MIPS(w/true ROIPS) suffers from high variance owing to its double propensity scores (when \( \alpha = 3.0 \)), although it does not introduce bias. Conversely, the MIPS introduces greater bias as \( \alpha \) increases, although Assumption 1 holds. This phenomenon is also discussed in our theoretical analysis in Section 3.1. the FM-IPS (DM) maintains the high bias owing to estimation errors, despite eliminating the position bias.

\section{CONCLUSION}
We first demonstrated that when both logging policy and position biases are present, the OPE estimators still have bias. We then proposed the MIPS(w/ROIPS), which leverages both the propensity scores of logging policies and reward observations to mitigate bias issues. Through experiments, we demonstrated that the MIPS (w/heuristic ROIPS) achieved the lowest MSE in real-world scenarios. 

However, when the assumption of no direct effect does not hold, a considerably strong assumption, our estimator may suffer from significant bias.

\section*{REFERENCES}
\printbibliography[heading=none]

\end{multicols}

\end{document}